\pdfoutput=1
\documentclass[11pt]{article}

\usepackage[final]{acl}

\usepackage{times}
\usepackage{latexsym}

\usepackage[T1]{fontenc}
\usepackage[utf8]{inputenc}

\usepackage{microtype}

\usepackage{inconsolata}

\usepackage{graphicx}

\usepackage{makecell} 
\usepackage{booktabs}    
\usepackage{pifont} 
\usepackage{xcolor}
\usepackage{amsmath}
\usepackage{multirow}
\usepackage{pifont}
\usepackage{colortbl}
\usepackage{float}
\usepackage{algorithm}
\usepackage{algorithmic}
\usepackage{tabularx}
\definecolor{Gainsboro}{rgb}{0.86, 0.86, 0.86}
\definecolor{darkgreen}{RGB}{0, 100, 0}
\usepackage{arydshln}
\usepackage[colorinlistoftodos]{todonotes}

\title{\texttt{ComplexFuncBench}: Exploring Multi-Step and Constrained Function Calling under Long-Context Scenario}

\author{
  Lucen Zhong$^{1}$ \quad Zhengxiao Du$^{1,2}$ \quad Xiaohan Zhang$^{1}$ \quad Haiyi Hu$^{1}$ \\
  \bf Jie Tang$^{2}$ \\
  $^1$Zhipu AI \quad $^2$Tsinghua University\\
  \texttt{lucen.zhong@aminer.cn}\quad \texttt{zx-du20@mails.tsinghua.edu.cn}
}

\begin{document}
\maketitle
\begin{abstract}
Enhancing large language models (LLMs) with real-time APIs can help generate more accurate and up-to-date responses. However, evaluating the function calling abilities of LLMs in real-world scenarios remains under-explored due to the complexity of data collection and evaluation. In this work, we introduce \texttt{ComplexFuncBench}, a benchmark for complex function calling across five real-world scenarios. Compared to existing benchmarks, \texttt{ComplexFuncBench} encompasses multi-step and constrained function calling, which requires long-parameter filing, parameter value reasoning, and 128k long context. Additionally, we propose an automatic framework, \texttt{ComplexEval}, for quantitatively evaluating complex function calling tasks. Through comprehensive experiments, we demonstrate the deficiencies of state-of-the-art LLMs in function calling and suggest future directions for optimizing these capabilities. The data and code are available at \url{https://github.com/THUDM/ComplexFuncBench}.
\end{abstract}

\section{Introduction}
Large Language Models (LLMs) achieve remarkable results on many general tasks \cite{hendrycksmeasuring, xu2023superclue, chen2021evaluating, cobbe2021training}. However, LLMs lack real-time and factual knowledge because they can only respond based on the information they were trained on \cite{zhao2023survey}. To address this limitation, recent research focuses on enhancing LLMs with function calling capabilities \cite{qin2024toollearningfoundationmodels, qin2023toolllm, chen2024facilitating, meetkai_functionary}. By integrating external tools and APIs, LLMs can deliver more accurate and up-to-date outputs.

% \cite{chen2024t, qiao2024benchmarking, nexusraven2023}
While many models \cite{openai_function_calling, anthropic_claude_3, team2024chatglm} are enhanced with function calling capabilities, evaluating function calling, especially complex calls, is still challenging. Researchers try various methods to evaluate the function calling capabilities of LLMs. Some works propose the prompt-based methods to assess the tool planning ability \cite{chen2024t, qiao2024benchmarking} of LLMs, but these do not assess the correctness of calling parameters. Meanwhile, some works evaluate the final state of the execution environments \cite{qin2023toolllm, lu2024toolsandbox}, yet ignore the correctness of the intermediate process. Moreover, some work uses rule-based matching methods\cite{berkeley-function-calling-leaderboard, wang2024mtu} to calculate the precision of function calls, but these methods are limited to simple scenarios and cannot effectively assess complex function calls. 

% To sum up, existing approaches only evaluate simple function calling abilities without fine-grained evaluation of the function calling process, while LLMs are supposed to conduct more complex function calls to respond to users in real-world scenarios. 

\begin{figure}[]
\centering 
\includegraphics[width=0.5\textwidth]{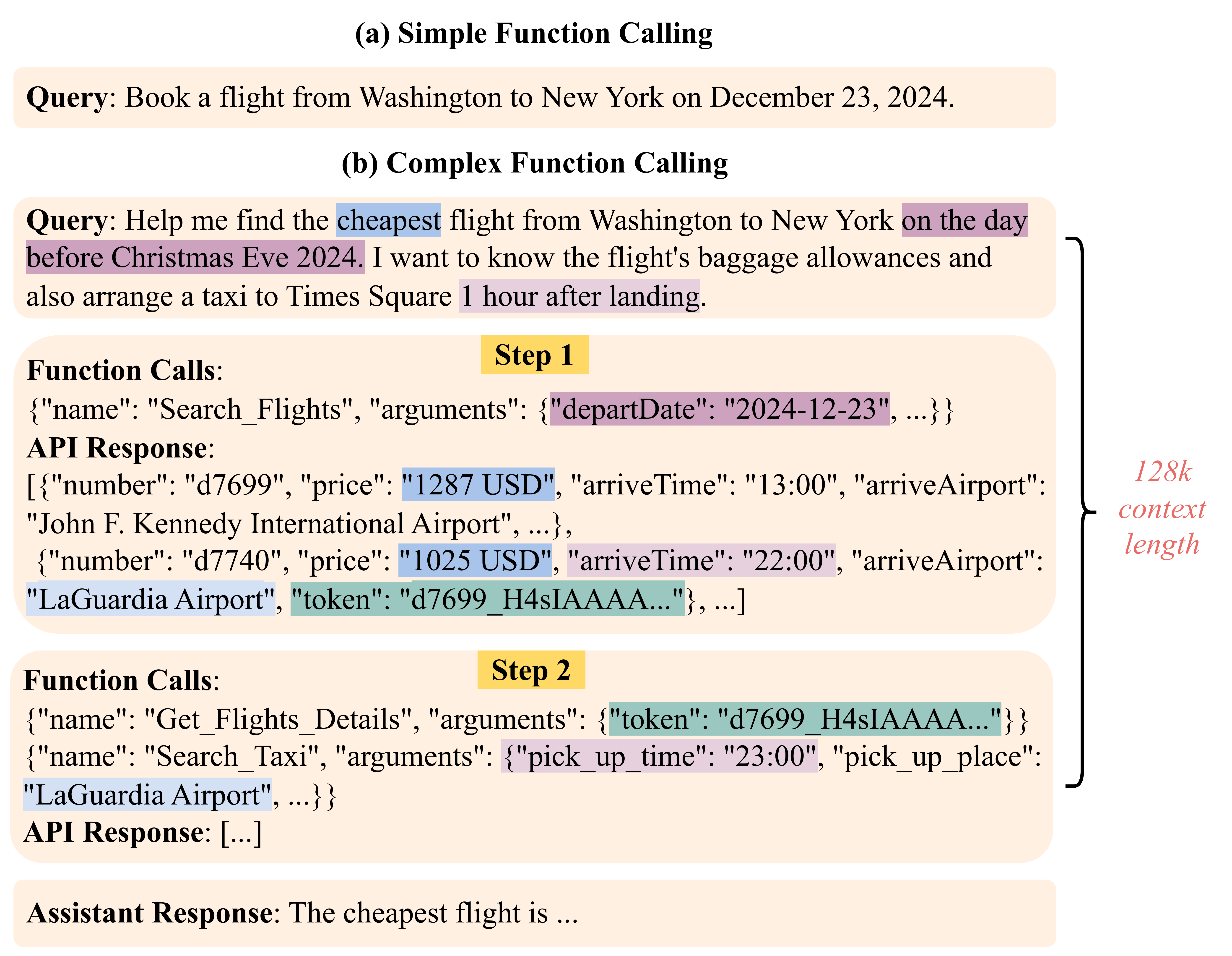}
\caption{(a) Simple Function Calling. (b) Complex Function Calling with \textcolor[rgb]{0.85, 0.701, 0.25}{multi-step}, \textcolor[rgb]{0.513, 0.619, 0.772}{constraints}, \textcolor[rgb]{0.654, 0.485, 0.595}{parameter value reasoning}, \textcolor[rgb]{0.415, 0.579, 0.552}{long parameter values} and \textcolor[rgb]{0.831, 0.380, 0.361}{long context}. Different colors correspond to the corresponding features marked in the figure.}
\label{intro_fig}
\vspace{-0.4cm}
\end{figure}

\begin{table*}[t]
    \centering
    \scalebox{0.7}{
    \begin{tabular}{ccccccc}
    \toprule
         & Real API Response & Multi-Step&  Constraints&  Parameter Value Reasoning &  Long Parameter Values& Long-Context\\
    \midrule
         \textbf{API-Bench}&  \textcolor{red}{\ding{55}}& \textcolor{red}{\ding{55}}&  \textcolor{red}{\ding{55}}&  \textcolor{red}{\ding{55}}&  \textcolor{red}{\ding{55}}& \textcolor{red}{\ding{55}}  \\
         \textbf{ToolBench}& \textcolor{green}{\ding{51}}&  \textcolor{green}{\ding{51}}&  \textcolor{red}{\ding{55}}&  \textcolor{red}{\ding{55}}&  \textcolor{red}{\ding{55}}& \textcolor{red}{\ding{55}}  \\
         \textbf{T-Eval}&  \textcolor{green}{\ding{51}}& \textcolor{green}{\ding{51}}&  \textcolor{red}{\ding{55}}&  \textcolor{red}{\ding{55}}&  \textcolor{red}{\ding{55}}& \textcolor{red}{\ding{55}}  \\
         \textbf{BFCL}& \textcolor{red}{\ding{55}}& \textcolor{green}{\ding{51}}&  \textcolor{red}{\ding{55}}&  \textcolor{red}{\ding{55}}&  \textcolor{green}{\ding{51}}& \textcolor{green}{\ding{51}} \\
         \textbf{Tool Sandbox}& \textcolor{red}{\ding{55}}&  \textcolor{green}{\ding{51}}&  \textcolor{red}{\ding{55}}&  \textcolor{red}{\ding{55}}&  \textcolor{red}{\ding{55}}& \textcolor{red}{\ding{55}}  \\
 \textbf{ComplexFuncBench}& \textcolor{green}{\ding{51}}& \textcolor{green}{\ding{51}}&  \textcolor{green}{\ding{51}}&  \textcolor{green}{\ding{51}}&  \textcolor{green}{\ding{51}}&\textcolor{green}{\ding{51}} \\
    \bottomrule
    \end{tabular}
}
    \caption{\texttt{ComplexFuncBench} compare with other function calling benchmarks.}
    \label{benchmark_comparison}
\vspace{-0.3cm}
\end{table*}

As shown in Figure \ref{intro_fig}, existing benchmarks directly provide parameters in the query and only require one-step function calling. In contrast, in real-world scenarios, users often implicitly provide parameter values rather than explicitly. LLMs are expected to infer the correct parameter values based on user constraints and API responses. Therefore, we define complex function calling from five aspects: (1) \textbf{Multi-step} in a single turn; (2) User-provided \textbf{constraints}; (3) \textbf{Parameter value reasoning} from implicit information; (4) \textbf{Long parameter values}; and (5) 128k \textbf{long context length}. 

The evaluation of complex function calling is currently underexplored due to several challenges. First, data annotation for complex function calling is time-consuming and laborious. Since current LLMs cannot do this task well, specially-trained senior annotators are required for annotation. Second, annotating a single valid path without disambiguating API response is challenging. Real-time API responses may contain multiple choices that satisfy user constraints at the current step, resulting in ambiguities in the parameters for subsequent calls. Third, traditional rule-based matching methods are inadequate for evaluating complex function calling since function calls with slightly different parameters can be equivalent. For example, parameter values can be expressed differently, such as \texttt{NY} and \texttt{New York}.
% parameters with default values can be omitted

% \footnote{\texttt{Hotels}, \texttt{Flights}, \texttt{Attractions}, \texttt{Car Rental} and \texttt{Taxi}} 
To address the above problems, we introduce \texttt{ComplexFuncBench}, a complex function calling benchmark with multi-step and constrained function calling under long-context scenario. We collect functions from the Booking.com \footnote{https://rapidapi.com/DataCrawler/api/booking-com15} website and manually annotate 1,000 complex function calling samples across five real-world domains. The annotation process is divided into three stages. At the coarse generation stage, we use GPT-4o to generate 1,000 queries along with the corresponding function calling paths to build a preliminary dataset to reduce the costs of human annotation. At the fine-grained annotation stage, we train senior annotators to select and label 100 complex samples with different function call paths based on the preliminary dataset to build a template dataset. Specifically, we first ensure that the annotated function calling path is the shortest one that fully addresses the query requirements while correcting any errors in parameter values. Then, we remove ambiguous information from API responses to ensure a single valid function calling path for each sample, avoiding parameter ambiguity during evaluation. At the generalization stage, we train junior annotators to generalize the template dataset from 100 to 1,000 samples by modifying specific information in the template queries to build \texttt{ComplexFuncBench}. Table \ref{benchmark_comparison} shows the difference between \texttt{ComplexFuncBench} and other function calling benchmarks.

We further propose an automatic evaluation framework \texttt{ComplexEval} for complex function calling evaluation. \texttt{ComplexEval} overcomes the limitations of traditional exact matching methods by using a multi-dimensional matching approach: 
(1)\ Rule-based matching treats two identical function calls as equivalent.
(2)\ Response-based matching treats function calls with the same API responses as equivalent.
(3)\ LLM-based matching prompts LLMs to decide the equivalence of function calls, allowing for different expressions of parameter values.
Additionally, \texttt{ComplexEval} provides error information as the API response to assess the self-correction ability of LLMs. Furthermore, we utilize LLMs to evaluate the completeness and correctness of the final model-generated responses.

\begin{figure*}[t] 
\centering 
\includegraphics[width=0.97\textwidth]{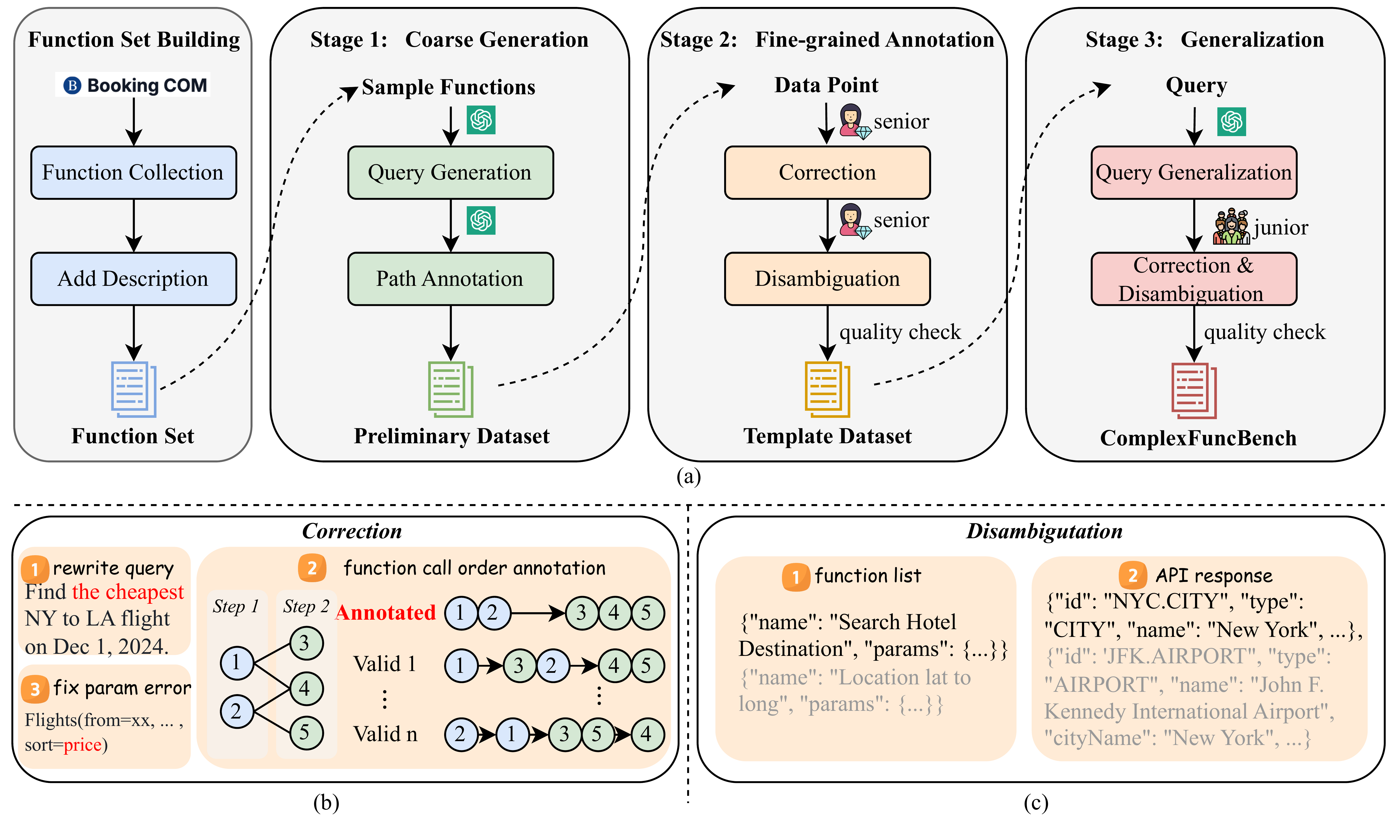}
\vspace{-0.1cm}
\caption{Overview of the data collection process. (a) is the high-level process of data collection. (b) is the example of human correction process.(c) is the example of disambiguation process. The \textcolor[rgb]{0.5, 0.5, 0.5}{grey} part is removed during annotation. A detailed annotation example is shown in Appendix \ref{annotation_example}.}
\label{data_collection}
\vspace{-0.2cm}
\end{figure*}

We conduct extensive experiments on various LLMs on \texttt{ComplexFuncBench}. We find that a significant portion of errors come from parameter value errors. Additionally, different models exhibit specific weaknesses in particular scenarios. Our contributions are as follows:
\begin{itemize}
\vspace{-0.1cm}
    \item We propose \texttt{ComplexFuncBench}, a dataset consisting of 1,000 samples with multi-step and constrained function calling that requires 128k long context.
\vspace{-0.1cm}
    \item We propose \texttt{ComplexEval}, an automatic evaluation framework for complex function calling. \texttt{ComplexEval} addresses previous evaluation challenges by integrating a multi-dimensional matching method.
\vspace{-0.1cm}
    \item The detailed analysis of the experiment results will guide future optimization of function calling capabilities.
\end{itemize}

\section{\texttt{ComplexFuncBench}}

\subsection{Data Collection}
% A function set exhibiting real-world scenarios is essential for constructing a dataset for complex function calling. Therefore, 
We collect daily-life APIs from Booking.com provided by RapidAPI to build our candidate function set, which includes 43 real-time APIs across five domains: \textit{Hotel}, \textit{Flight}, \textit{Attraction}, \textit{Car Rental}, and \textit{Taxi}. We find that the function descriptions from the website are incomplete and even incorrect. Therefore, we manually correct and annotate the detailed function description for each function to build a high-quality function set.

The \texttt{ComplexFuncBench} dataset collection has three stages: coarse generation, fine-grained annotation, and generalization. The overview of the data collection process is shown in Figure \ref{data_collection}.

\subsubsection{Stage 1: Coarse Generation}
Annotating complex function calling samples is a multifaceted process that includes query construction, parameter annotation, function call order arrangement, and API response disambiguation. To mitigate the costs of human annotation, we leverage the function calling ability of GPT-4o to generate 1,000 preliminary complex function calling samples as a reference to assist human annotators.

Specifically, we randomly sampled several functions from the function set and specially designed a prompt for GPT-4o to generate 1,000 queries requiring complex function calling across different domains. The detailed prompt is shown in Appendix \ref{appendix:query_generation}. We leverage GPT-4o's function calling interface to automatically annotate these queries with function calling results to construct a preliminary dataset. However, the automatically annotated data exhibits severe errors due to current models' limited complex function calling capabilities, such as parameter value and function call order errors. 

\subsubsection{Stage 2: Fine-grained Annotation}
To ensure the precision and reliability of the benchmark dataset, we propose a fine-grained annotation process to refine the preliminary dataset. Many automatically generated samples in the preliminary dataset failed at the first step, making subsequent human annotation difficult. Therefore, we engage human annotators to select 100 samples with relatively complete function calling paths from the preliminary dataset, encompassing various function calling steps and different function calling paths. A senior annotator will annotate the selected samples through the following steps.

% Path and parameter correction
\paragraph{Correction} Annotators must correct the GPT-generated samples from three aspects. First, rewrite the query with detailed information, like the \textsc{arrive\_time} of the taxi and the \textsc{check\_in\_date} of the hotel. Second, annotate the function call order. Specifically, they add missing function calls and remove unnecessary ones to get the whole function call path that can comprehensively address the query requirements. Then, they adjust the function call order to build the shortest function call path as the final annotated path. As shown in Figure \ref{data_collection}(b), a query may have multiple valid calling paths to complete the task. We annotate the shortest path for quantitative evaluation later. Third, correct function calling parameter errors, such as parameter type mismatches, incorrect parameter values, parameter hallucinations, and missing parameters. 

\paragraph{Disambiguation} After correction, human annotators must ensure that the functions and the calling parameters are unambiguous. First, overlapping functionalities are removed from the function list for each sample to eliminate ambiguity in functions. For example, obtaining the latitude and longitude of a city can be achieved through both \textsc{Search\_Hotel\_Destination} and \textsc{Location\_lat\_to\_long} function. Therefore, these two functions will not be available in the same sample. Second, to disambiguate the calling parameter values, we will delete API responses that may cause parameter ambiguity for subsequent function calls. As shown in Figure \ref{data_collection}(c), when searching hotels in New York City, the initial step might return multiple location IDs of NY(e.g., for downtown, airports, attractions, etc.). Only one of these location IDs is retained during the annotation process to prevent ambiguity in subsequent calls. Note that there is no need to delete results unrelated to New York City, as those do not introduce ambiguity.

% An annotation example is shown in Appendix \ref{annotation_example}. 
We perform quality checks on all 100 samples until there are no errors. Finally, we get 100 complex function calling paths, including single-domain and cross-domains.

\subsubsection{Stage 3: Generalization}
To reduce the variance of the evaluation results, we generalize each sample in the template dataset to 10 samples. This expansion increases the \texttt{ComplexFuncBench} dataset size from 100 to 1,000. By building a larger dataset, we can get more reliable and stable results from the benchmark.

Specifically, we use the 100 queries in the template dataset as examples and design prompts for GPT-4o to modify elements such as locations and constraints, generating nine new queries for each original query. The detailed prompt is shown in Appendix \ref{appendix:generalization}. These 10 queries share the same function calling path but differ in parameter values for each call. We then trained junior annotators to fill in the respective function call parameter values based on the given path, delete API responses to avoid disambiguation, and obtain a new complex function calling sample. The annotation difficulty and human effort required for generalization are significantly lower than the cost of fine-grained annotation, enabling several junior annotators to complete large-scale annotations effectively.

\subsection{Data Composition}
Finally, we obtained 1,000 complex function calling samples, which comprises 600 single-domain samples, with 150 samples each from the \texttt{Hotels}, \texttt{Flights}, \texttt{Car Rental}, and \texttt{Attraction} domains, and 400 cross-domain samples. The \texttt{taxi} domain only has two functions, so it is only used for cross-domain. The average number of function calling steps in \texttt{ComplexFuncBench} is 3.26, while the average number of function calls per sample is 5.07. Detailed statistics are shown in Appendix \ref{appendix:data_stat}.

\begin{figure*}[] %H为当前位置，!htb为忽略美学标准，htbp为浮动图形
\centering %图片居中
\includegraphics[width=0.97\textwidth]{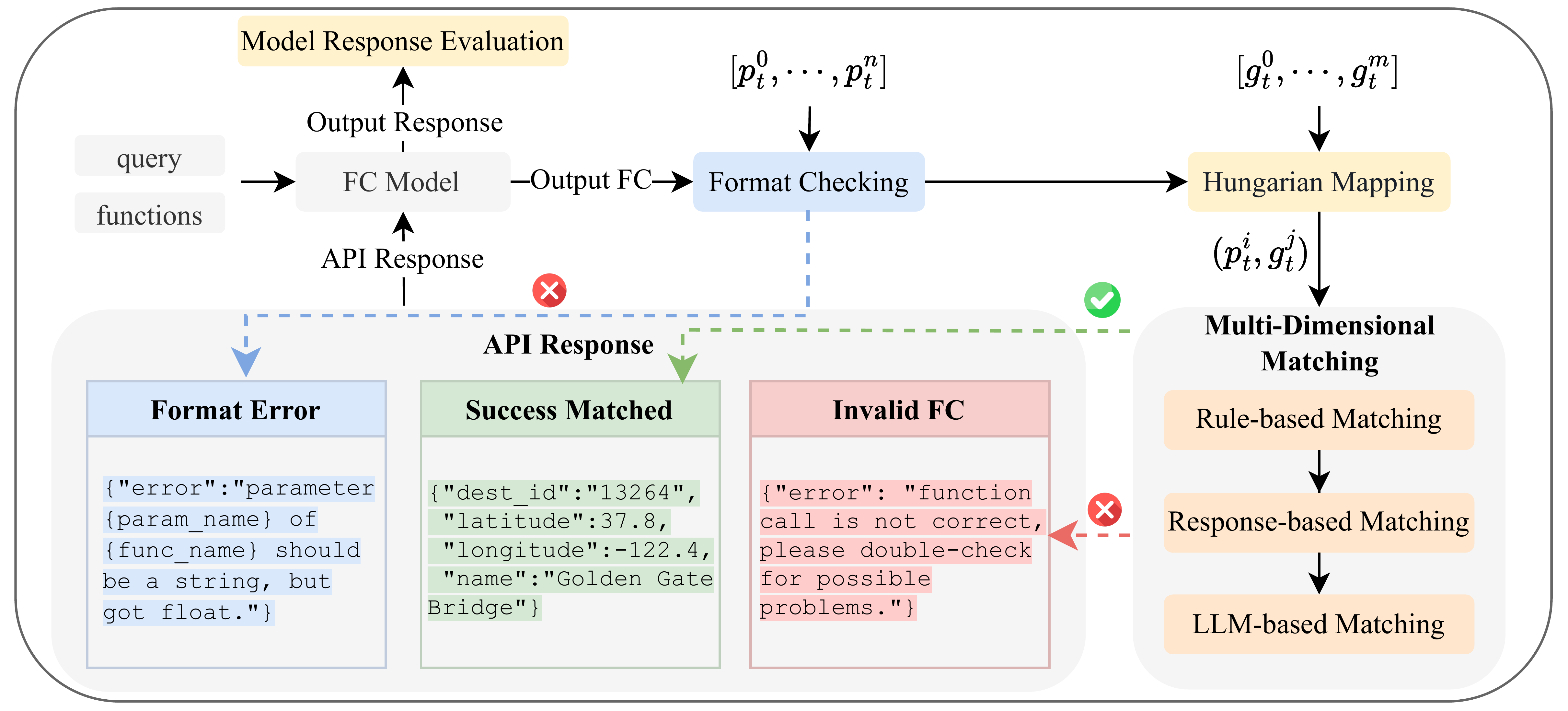}
%插入图片，[]中设置图片大小，{}中是图片文件名
\vspace{-0.1cm}
\caption{Overview of \texttt{ComplexEval}. Different colors represent different API response types. Color \textcolor{blue}{blue} represents format error with specific error message. Color \textcolor[rgb]{0.0, 0.5, 0.0}{green} represents correct function call with corresponding golden API response. Color \textcolor{red}{red} represents invalid function call with general error message.}
\label{auto_evaluation}
\vspace{-0.2cm}
\end{figure*}

\section{\texttt{ComplexEval}: Automatic Evaluation}
We propose \texttt{ComplexEval} to quantitatively evaluate models' complex function calling ability and response generation ability. Figure \ref{auto_evaluation} is the overview of \texttt{ComplexEval}.

\subsection{Function Calling Evaluation}
\label{fc_eval}
Given the user query $q$ and the available function list $f$, the model continually generates the next-step function calls until the final response is generated.

Specifically, at step $t$, the model generates a list consisting of $n$ predicted function calls, denote as $[p_{t}^{0}, \cdots, p_{t}^{n}]$. And the golden function call list contains $m$ function calls $[g_{t}^{0}, \cdots, g_{t}^{m}]$. We aim to evaluate the correctness of each predicted function call $p_{t}^{i}$ against the golden function call list. We will append the corresponding API response to the dialogue history for each $p_{t}^{i}$. Based on the updated dialogue history, the model then predicts the next-step function calls. The evaluation process includes format checking, Hungarian mapping for function calls, and multi-dimensional matching. 
 
\paragraph{Format Checking} First, we check the format of each $p_{t}^{i}$ in the predicted function call list. The format check includes three parts: (1) verifying that the called function is in the available function list, (2) ensuring that all required parameters are included in the call, and (3) confirming that the type of each parameter meets the requirement in the function description. If $p_{t}^{i}$ fails the format checking, we use a specific error message as the API response. This approach allows the model to undergo a trial-and-error process, and we can evaluate the model's self-correction ability.

\paragraph{Hungarian Mapping} For remaining predicted function calls with correct formats, we use bge-large-en-v1.5 model\cite{bge_embedding} to obtain the text embedding for each function call. Specifically, we concatenate the function name and parameters into a single string for text encoding. The predicted and golden function calls list embedding are denoted as $\mathbf{e}_p$ and $\mathbf{e}_g$, respectively. Then, we calculate the cosine similarity between $\mathbf{e}_p$ and $\mathbf{e}_g$ and employ the Hungarian mapping algorithm to get a mapping list, e.g., $(p_t^i, g_t^j)$ denotes the $i_{th}$ predict function call mapping with the $j_{th}$ golden function call in at step $t$.
\begin{equation}
(i, j) = \mathrm{HungarianMatch}(-\mathbf{e}_p \times \mathbf{e}_{g}^T)
\label{match}
\end{equation}

% Add examples for different matching metheds, see appendix.
\paragraph{Multi-Dimensional Matching} Based on the mapping list, we propose a multi-dimensional matching method to decide the equivalence between $p_t^i$ and $g_t^j$.
\begin{itemize}
    \item \textit{Rule-based Matching} determines equivalence through exact matching. Two identical function calls are considered equivalent.
    \item \textit{Response-based Matching} determines equivalence by comparing the returned API response of the two function calls. Two function calls are considered equivalent when the returned results are identical. For example, if parameter \texttt{adults} default value is 1, then \textit{Search\_Hotel(New\_York, adults=1)} and \textit{Search\_Hotel(New\_York)} are equivalent.
    \item \textit{LLM-based Matching} leverages GPT-4o to determine the equivalence. Specifically, Parameter values can be expressed in different forms as long as the meaning is the same. For example, \textit{New York} and \textit{NY}, \textit{Narita International Airport} and \textit{Tokyo Narita International Airport} are equivalent. Detailed prompt is shown in Appendix \ref{subsec:llm-based-match}.
\end{itemize}

If $p_t^i$ is equivalent to $g_t^j$, the annotated API response of  $g_t^j$ is used as the API response of $p_t^i$. If $p_t^i$ is not equivalent to $g_t^j$, we use a fixed system error message as the API response of $p_t^i$. We will update the golden function call list by removing successfully matched calls and adding the next-step calls, as shown in Appendix \ref{golden_updating}. We continually call the model to generate the next-step function calls until the final model response is generated.

\paragraph{Metrics} We use \texttt{Success Rate} and \texttt{Call Accuracy} as metrics. \texttt{Success Rate} measures the overall task completion by calculating the proportion of samples that successfully complete the task.  \texttt{Call Accuracy} calculates the proportion of correct function calls as in equation (\ref{call_acc}).
\begin{equation}
\text { Call Acc }=\frac{\sum_{i=1}^N c_i}{\sum_{i=1}^N n_i}
\label{call_acc}
\end{equation}
where $N$ is the number of samples, $n_i$ is the total number of function calls in sample $i$, and $c_i$ is the number of correct function calls in sample $i$.

\subsection{Model Response Evaluation}
After obtaining the model's final response, we use GPT-4o to further evaluate the response quality from two aspects: completeness and correctness. The detailed prompt is shown in Appendix \ref{response_eval_prompt}
\paragraph{Completeness} Whether the response comprehensively addresses all the requirements proposed by the user. Given the query $q$ and generated response $r_{gen}$, GPT-4o will give a $\text{score} \in \{0, 1, 2\}$ as the completeness score. $\text{score} = 0$ indicates none of the user requirements were addressed, $\text{score}=1$ indicates partial fulfillment and $\text{score}=2$ indicates full fulfillment. We calculate the average score of all samples as the final completeness score.

\paragraph{Correctness} Whether the provided answers are accurate and align with the API response. Given the complete dialogue history $h$ and generated response $r_{gen}$, GPT-4o will give a $\text{score} \in \{0, 1, 2\}$ as the correctness score. $\text{score} = 0$ indicates totally incorrect, $\text{score}=1$ indicates partially correct and $\text{score}=2$ indicates totally correct. We calculate the average score of all samples as the final correctness score. 

\section{Experiments}

\subsection{Settings}
We select function calling models with 128k context length, encompassing 12 models from 6 institutions, to comprehensively investigate current model performance on \texttt{ComplexFuncBench}. Concretely, for closed-source models, we evaluate the latest version of GPT-4o and GPT-4-Turbo\cite{OpenAIModels}, Claude 3.5 Sonnet and Haiku\cite{anthropic_claude_3} and GLM-4-Long\cite{team2024chatglm}. For open-source models, we evaluate Llama-3.1 series from 8B to 405B\cite{dubey2024llama}, Qwen2.5\cite{qwen2.5} 7B and 72B, and Mistral Large 2\cite{mistral}.

\subsection{Main Results}
\begin{table*}[htbp]
    \centering
    \scalebox{0.55}{
    \begin{tabular}{ccccccccccccccc}
         \toprule
         \multirow{3}{*}{\textbf{Model}} &  \multicolumn{2}{c}{Hotels}&  \multicolumn{2}{c}{Flights}&  \multicolumn{2}{c}{Car Rental}&  \multicolumn{2}{c}{Attraction}&  \multicolumn{2}{c}{Cross}& \multicolumn{2}{c}{Overall} & \multirow{2}{*}{Completeness} & \multirow{2}{*}{Correctness} \\
         \cmidrule{2-13}
         &  Success&  Call Acc &  Success&  Call Acc&  Success&  Call Acc&  Success&  Call Acc&  Success& Call Acc& Success&Call Acc \\
         \midrule
         \rowcolor[gray]{0.95} \multicolumn{15}{c}{\textit{close-source models}}\\
         \midrule
         Claude-3.5-Haiku&  36.00&  50.62&  50.67&  75.63&  59.33&  74.05&  58.00&  75.37&  38.00& 70.00& 45.80&69.50 & \underline{1.79} & 1.71\\
         Claude-3.5-Sonnet&  54.67&  68.17&  54.00&  79.50&  \underline{76.67}&  \underline{86.01}&  69.33&  \underline{83.33}&  \textbf{57.00}& \textbf{79.33}& \textbf{61.00}&\underline{79.27} & \textbf{1.84} & \textbf{1.85}\\
         GLM-4-Long& \underline{56.00}& 63.98& \textbf{66.67} & \underline{84.38}& \textbf{77.33}& 85.71& \underline{72.67}& \underline{83.33}& 40.50& 72.75& 57.10& 76.35&  1.72 & 1.74   \\
         GPT-4-Turbo& 54.67& \underline{68.48}& 48.67& 76.5& 44.67& 71.14& 70.67& 76.48& 41.75& 69.38& 49.50& 71.38&  1.72& \underline{1.81} \\
         GPT-4o&  \textbf{70.00}&  \textbf{81.99}&  \underline{65.33}&  \textbf{85.50}&  72.00&  \textbf{86.88}&  \textbf{82.00}&  \textbf{87.59}&  \underline{42.75}& \underline{75.13}& \underline{60.50}&\textbf{80.55} &  1.66& 1.75\\
         \midrule
         \rowcolor[gray]{0.95} \multicolumn{15}{c}{\textit{open-source models}}\\
         \midrule
         Qwen2.5-7B&  \underline{2.00}&  \underline{20.65}&  0.00&  \underline{5.13}&  \underline{4.67}&  \underline{6.41}&  \textbf{14.67}&  \textbf{35.18}&  \textbf{4.5}& \underline{21.41}& \underline{5.0}&\underline{18.19} & \textbf{1.5} & \textbf{1.47} \\
 Llama-3.1-8B& 0.00& 0.00& 0.00& 1.00& 0.00& 1.89& 0.67& 2.78& 0.00& 1.00& 0.10&1.34 & 0.18 & 0.09 \\
 GLM-4-9B& \textbf{19.33}& \textbf{31.52}& \textbf{11.33}& \textbf{34.00}& \textbf{16.0}& \textbf{25.36}& \underline{10.67}& \underline{29.26}& \underline{2.00}& \textbf{25.46}& \textbf{9.40} & \textbf{27.97} & \underline{1.15} & \underline{1.03} \\
 \midrule
 Llama-3.1-70B& 2.00& 10.71& 0.67& 2.63& 6.67& 10.06& 4.67& 11.11& 1.50& 8.13& 2.70&8.17 & 0.67 & 0.36\\
 Llama-3.1-405B&  3.33& 13.51& 2.66& 10.75& 4.00& 15.74& 14.00& 18.52& 1.00& 9.21& 4.00&11.87 & 0.43 & 0.30 \\
 Qwen2.5-72B& \textbf{40.00}& \textbf{60.24}& \textbf{31.33}& \underline{49.25}& \textbf{48.67}& \underline{57.58}& \textbf{63.33}& \textbf{67.41}& \textbf{31.50}& \textbf{59.00}& \textbf{40.10}&\textbf{58.32}&\textbf{1.80} & \textbf{1.75}\\
 Mistral Large 2& \underline{19.33}& \underline{34.32}& \underline{20.67}& \textbf{52.88}& \underline{40.67}& \textbf{58.16}& \underline{25.33}& \underline{40.18}& \underline{10.50}& \underline{50.54}& \underline{20.10}&\underline{48.78}& \underline{0.94} & \underline{1.0} \\
    \bottomrule
    \end{tabular}
}
    \caption{\textbf{Main Results}. We categorize models as close-source, open-source under 10B, and open-source above 10B. Top two results for each category are highlighted in \textbf{bold} and \underline{underline}. The specific endpoint of open-source models are: \texttt{gpt-4o-2024-08-06}, \texttt{gpt-4-turbo-2024-04-09}, \texttt{claude-3-5-sonnet-20241022} and \texttt{claude-3-5-haiku-20241022}. }
    \label{tab:main_result}
\end{table*}

The main results of function call evaluation and model response evaluation are shown in Table \ref{tab:main_result}.

\paragraph{Function Call Evaluation}  The complex function calling ability of the closed-source models is superior to that of the open-source models. Claude-3.5-Sonnet, GPT-4o, and GLM-4-Long perform comparably among the closed-source models, with overall success rates of 61.0\%, 60.5\%, and 57.1\%, respectively. Among the open-source models, Qwen2.5-72B performs the best, achieving a task success rate of 40.1\%. The Llama3.1 series models do not possess the ability for complex function calling; the experiment results indicate that Llama3.1 suffers from disordered function call sequences and severe parameter hallucinations during the function calling process. Moreover, models with less than 10B parameters cannot handle the complex function calling well; the best-performing model, GLM-4-9B, only achieved a success rate of 8.4\%.

Different models' performance varies in different domains due to the distinct challenges presented by each domain. Specifically, GPT-4o achieved notable success rates of 70\% and 82\% in the Hotel and Attraction domains, respectively, significantly outperforming other models. These domains require a model to comprehensively understand the function descriptions to infer the function calling order. Conversely, Claude-3.5-Sonnet handles more complex function calls in cross-domain scenarios better. Meanwhile, GLM-4-Long excelled in the Flights and Car Rental domains, which require handing long parameter values approaching 600 characters.

\paragraph{Model Response Evaluation} For closed-source models, Claude-3.5-Sonnet outperforms other models in both the completeness and correctness of responses, achieving scores of 1.84 and 1.85, respectively. However, despite GPT-4o's high success rate, it shows the least complete performance with a score of 1.66. For the open-source model, Qwen2.5-72B generates the most complete and correct response, with completeness and correctness scores of 1.80 and 1.75, which is superior to other open-source models.

\subsection{Results Analysis}
To compare the performance between different models more precisely, we select the top three closed-source models and the leading open-source model for in-depth analysis.
\subsubsection{Error Analysis}

\begin{figure}[htbp] %H为当前位置，!htb为忽略美学标准，htbp为浮动图形
\centering %图片居中
\includegraphics[width=0.45\textwidth]{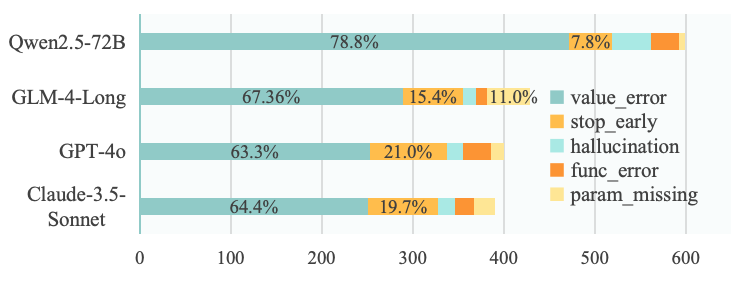}
%插入图片，[]中设置图片大小，{}中是图片文件名
\vspace{-0.1cm}
\caption{Error type analysis for different models.}
\label{error_distribution}
\vspace{-0.2cm}
\end{figure}

We classify the errors encountered during the complex function calling process into five types: \texttt{func\_error}, \texttt{param\_missing}, \texttt{hallucination}, \texttt{value\_error}, and \texttt{stop\_early}. \texttt{func\_error} indicates calling a wrong or invalid function. \texttt{param\_missing} refers to the omission of crucial parameters. \texttt{hallucination} refers to generating random parameters not provided by the user. \texttt{value\_error} denotes parameter values errors. \texttt{stop\_early} means generating the final response without calling all required functions.

Figure \ref{error_distribution} presents the error analysis results for four models. We can see that \texttt{value\_error} accounts for a significant portion of errors in all models, particularly for Qwen2.5-72B, which has a \texttt{value\_error} rate of 78.8\%. The high \texttt{value\_error} rate indicates that \texttt{ComplexFuncBench} poses significant challenges for LLMs in constrained parameter value reasoning and long-context parameter extraction. In addition to \texttt{value\_error}, all four models tend to stop function calling without gathering complete information. This issue is especially severe for Claude-3.5-Sonnet and GPT-4o, with \texttt{stop\_early} rates of 19.7\% and 21.0\%, respectively. Moreover, GLM-4-Long suffers from missing parameters with a \texttt{param\_missing} rate of 11.0\%, while other models show fewer parameter missing problems. Although less common compared to \texttt{value\_error}, \texttt{func\_error} still occurs, showing the model lacks understanding of function documentation.

\begin{figure*}[htbp] %H为当前位置，!htb为忽略美学标准，htbp为浮动图形
\centering %图片居中
\includegraphics[width=0.85\textwidth]{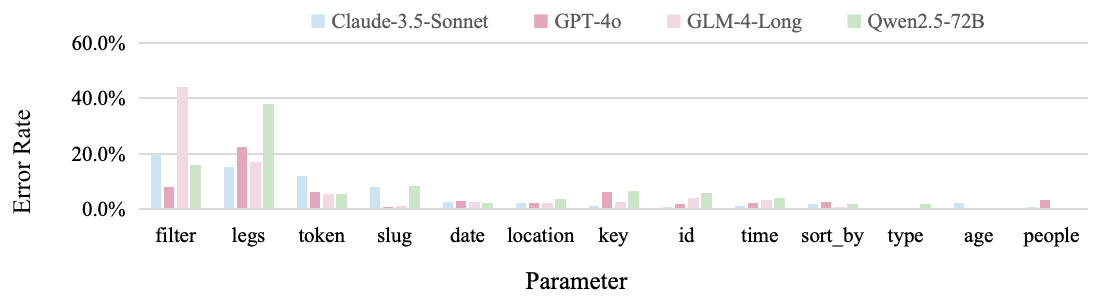}
%插入图片，[]中设置图片大小，{}中是图片文件名
\vspace{-0.1cm}
\caption{Error rates for each parameter type of different models}
\label{value_error}
\vspace{-0.2cm}
\end{figure*}

We analyze the error rate for each parameter type to further investigate the causes of the \texttt{value\_error} in different models. Specifically, we divide all parameters into different types, such as date-related, time-related, and location-related parameters. Appendix \ref{appendix:param_type} gives examples of different parameter types. Then, we calculate the error rates of different models on different parameter types. The result is shown in Figure \ref{value_error}. 

Although slight variations exist in the error distribution for different models, the incorrect parameters related to $filter$ and $legs$ are relatively higher than other parameters. The $filter$ parameter challenges the model to infer function calling orders from the parameter description. GLM-4-Long fails more than 40\% on this parameter. The $legs$ parameter needs the model to gather multiple information from the user query and API response. Qwen2.5-72B fails 38.1\% on this parameter. Claude-3.5-Sonnet cannot understand the constrained user query and consistently retrieves the wrong $token$ parameter. Claude-3.5-Sonnet and GPT-4o cannot understand the relationships between parameters specified in the function description and often retrieve the wrong $slug$ parameter from the API response. The results indicate that it is challenging for models to complete complex function calling with constraints and long context, and there is still room for improvement.

\subsubsection{Function Calling Steps Analysis}

\begin{figure}[htbp] %H为当前位置，!htb为忽略美学标准，htbp为浮动图形
\centering %图片居中
\includegraphics[width=0.48\textwidth]{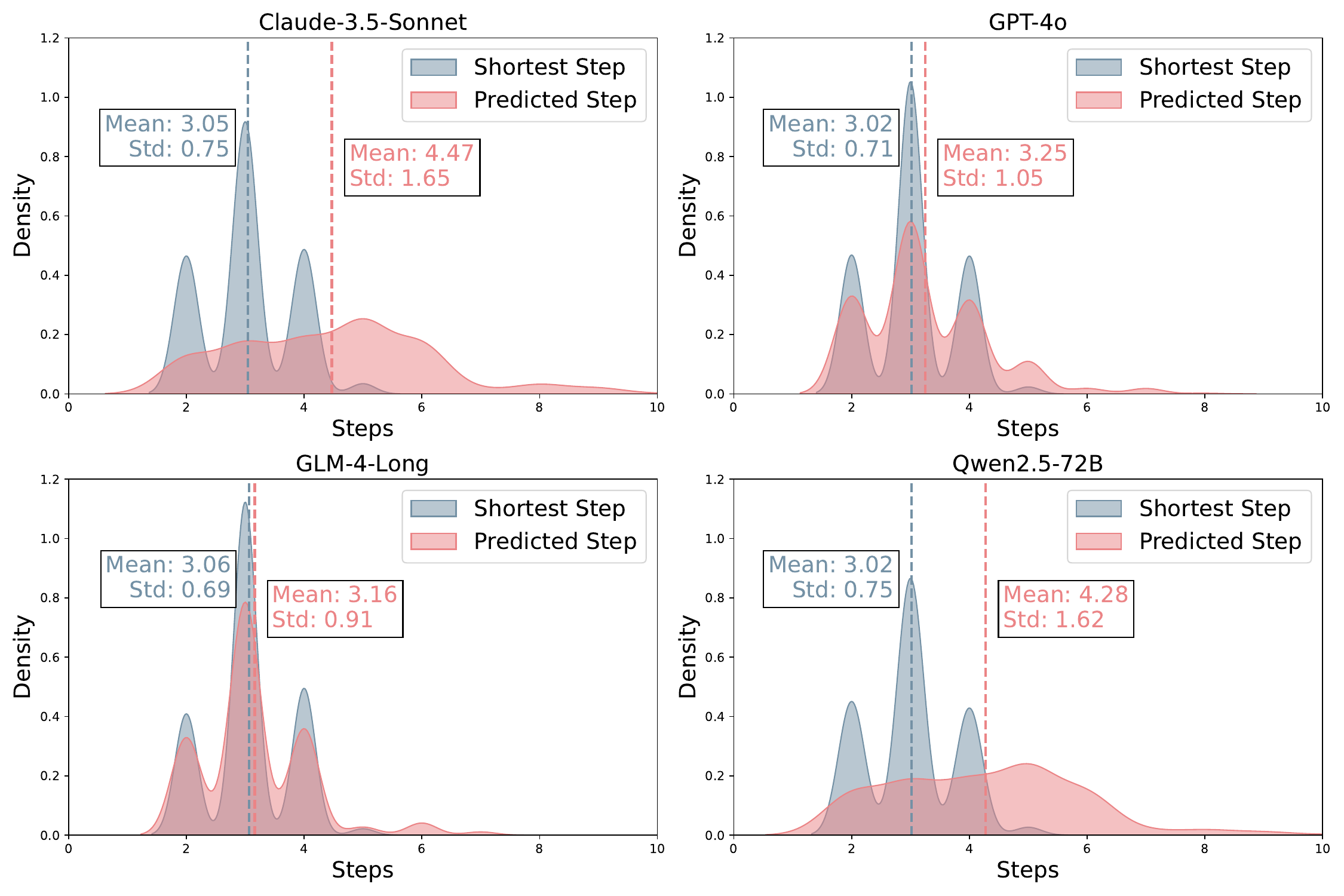}
%插入图片，[]中设置图片大小，{}中是图片文件名
\vspace{-0.5cm}
\caption{Function calling steps distribution.}
\label{length_distribution}
\vspace{-0.2cm}
\end{figure}

Figure \ref{length_distribution} compares the annotated shortest calling steps with the predicted calling steps among different models. We can see that the predicted calling steps often exceed the shortest calling steps due to the models' distinct planning strategies. Specifically, Claude-3.5-Sonnet and Qwen2.5-72B require an average of 4.47 and 4.28 steps, respectively, beyond the shortest steps. These models generally need 1–3 extra steps more than the shortest steps. As the task complexity increases, these models need more than 10 steps for task completion. In contrast, models such as GLM-4-Long and GPT-4o tend to complete tasks with fewer steps. The GLM-4-Long's average calling step is only 0.13 steps longer than the shortest steps.

\section{Related Work}
\paragraph{Function Calling with LLMs} Numerous models have been developed to enable function calling within Large Language Models (LLMs) to address challenges in real-world applications. Some foundation models \cite{openai_function_calling, team2024chatglm, anthropic_claude_3} are inherently equipped with native function calling capabilities, allowing seamless integration of API functionalities. In addition to these, some work proposes to fine-tune base models with specialized function calling datasets \cite{qin2023toolllm, patil2023gorilla, gao2024confucius, meetkai_functionary, liu2024toolace}. By training models on curated datasets, these works enable the models to dynamically generate function calls and execute functions, bridging the gap between language model generation and real-world scenarios. 

\paragraph{Function Calling Evaluation} A reliable and comprehensive function calling benchmark is crucial for providing optimization directions for function calling models. Several works propose to use llm-based methods to evaluate the tool planning capabilities of LLMs \cite{qiao2024benchmarking, nexusraven2023}. However, these methods can not capture the correctness of the function call parameters since they generate natural language thoughts instead of formatted function calls. Therefore, some studies propose rule-based matching methods, such as AST tree analysis, to evaluate the precision of function calling \cite{berkeley-function-calling-leaderboard, li2023comprehensive, wang2024mtu, chen2024t}. However, these studies only evaluate single function calling without assessing complex function calling scenarios. Furthermore, some work proposes to use state-based \cite{qin2023toolllm, lu2024toolsandbox, berkeley-function-calling-leaderboard} methods to evaluate the final state of the system without considering the intermediate function call procedure, resulting in the randomness of the evaluation results. While previous research lacks a thorough evaluation of complex function calling in real-world scenarios, this work introduces a benchmark for function calling in more complex settings.

\section{Conclusion}
In this work, we introduce \texttt{ComplexFuncBench}, a benchmark dataset consisting of 1,000 samples from five realistic scenarios, and an automatic evaluation framework \texttt{ComplexEval} for complex function calling. Unlike existing benchmarks, \texttt{ComplexFuncBench} features multi-step and constrained function calling with parameter value reasoning, long-context parameter extraction, and long parameter value filling. We show the deficiency of current state-of-the-art function calling models through comprehensive experiments. We hope \texttt{ComplexFuncBench} can guide the research community in developing function calling capabilities of foundation models.

% \section*{Acknowledgments}

% Bibliography entries for the entire Anthology, followed by custom entries
%\bibliography{anthology,custom}
% Custom bibliography entries only
\bibliography{custom}

\appendix

\section{Data Annotation}

\subsection{An Example of Annotation Process}
Table \ref{tab:annotate_example} gives a detailed example of the annotation process.
\label{annotation_example}

\subsection{Data Statistics}
\label{appendix:data_stat}
The data statistics for \texttt{ComplexFuncBench} are shown in Table \ref{tab:data_stat}.

\begin{table}[h]
    \centering
    \scalebox{0.65}{
    \begin{tabular}{lcccccc}
    \toprule
         &  Hotels&  Flights&  Car Rental&  Attraction& Cross&Total\\
    \midrule
         \# Samples&  150&  150&  150&  150& 400&1000\\
         Avg. Steps&  3.33&  3.33&  2.87&  2.86& 3.5&3.26\\
         Avg. Calls&  4.29&  5.33&  4.57&  3.6& 6.0&5.07\\
    \bottomrule
    \end{tabular}
    }
    \caption{Data statistics for \texttt{ComplexFuncBench.}}
    \label{tab:data_stat}
\end{table}

\subsection{Prompt for Query Generation.}
\label{appendix:query_generation}
\begin{figure*} %H为当前位置，!htb为忽略美学标准，htbp为浮动图形
\centering %图片居中
\includegraphics[width=0.9\textwidth]{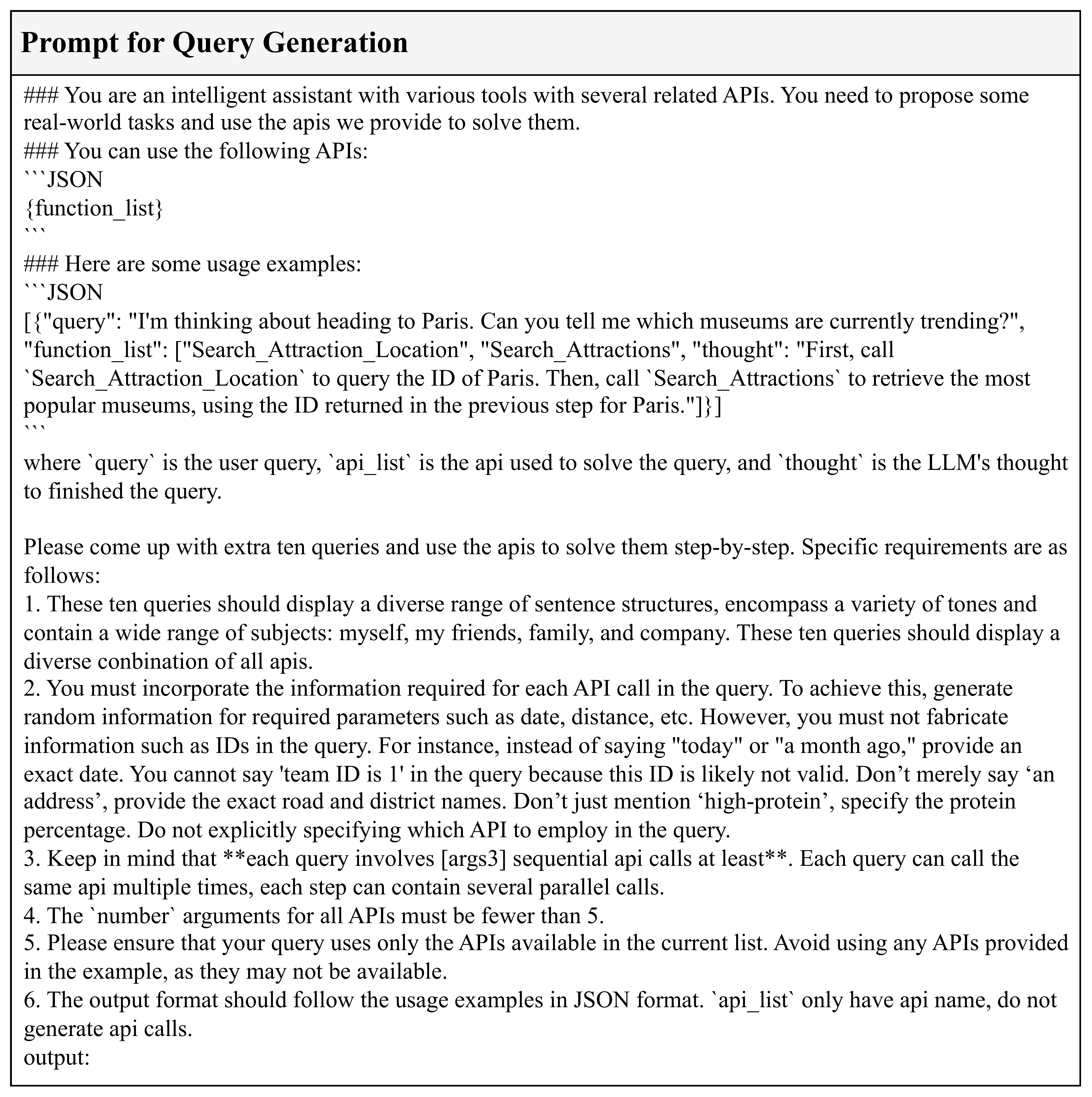}
%插入图片，[]中设置图片大小，{}中是图片文件名
\vspace{-0.1cm}
\caption{Prompt for Query Generation.}
\label{query_generation}
\vspace{-0.2cm}
\end{figure*}

Prompt for query generation is shown in Figure \ref{query_generation}.

\subsection{Prompt for Query Generalization}
\label{appendix:generalization}
\begin{figure*} %H为当前位置，!htb为忽略美学标准，htbp为浮动图形
\centering %图片居中
\includegraphics[width=0.9\textwidth]{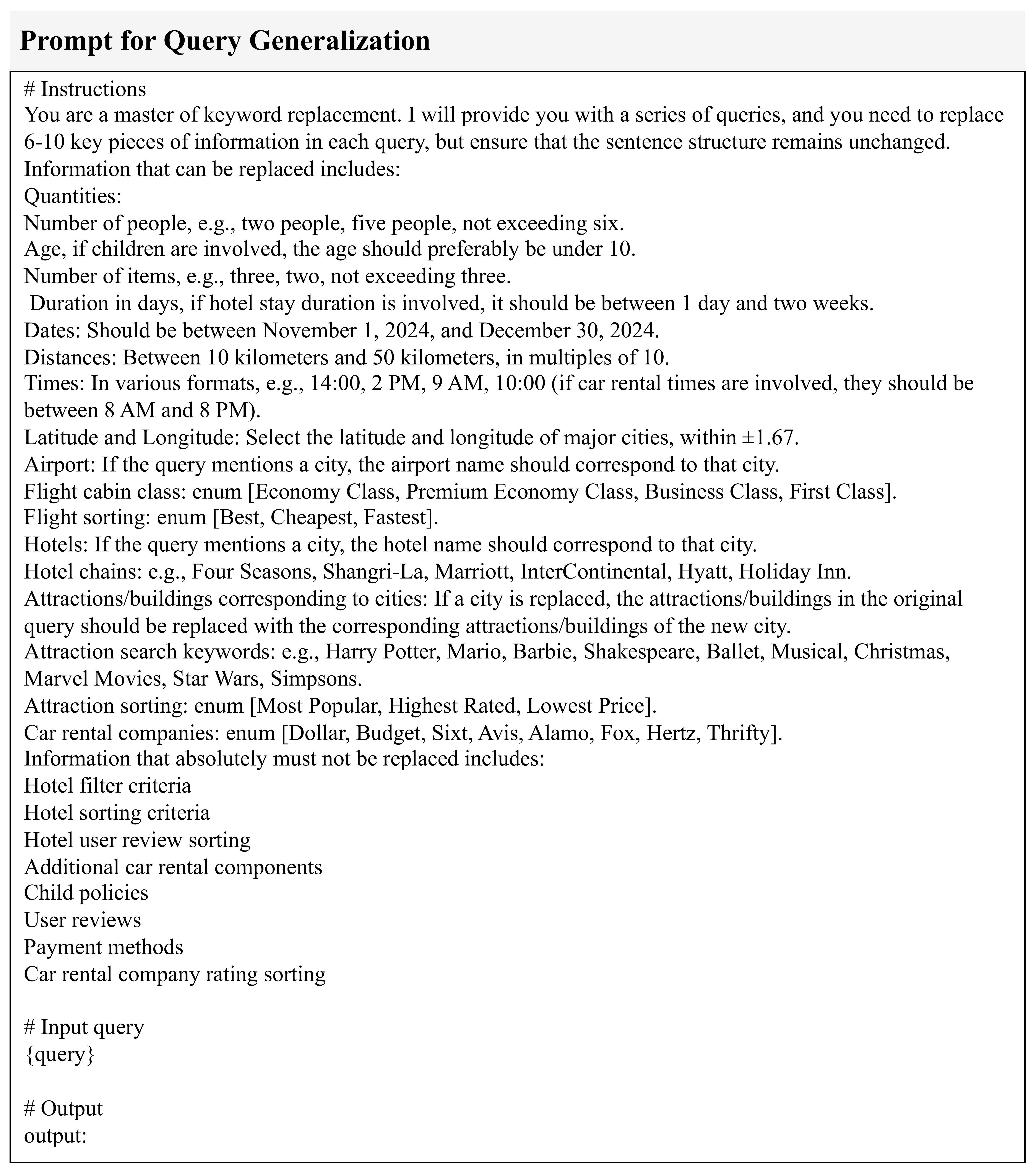}
%插入图片，[]中设置图片大小，{}中是图片文件名
\vspace{-0.1cm}
\caption{Prompt for Query Generalization.}
\label{query_generalization}
\vspace{-0.2cm}
\end{figure*}

Prompt for query generalization at the generalization stage is shown in Figure \ref{query_generalization}.

\section{Automatic Evaluation}

\subsection{Experiment Setting}
For a fair comparison of the LLM's complex function calling, we use the same generation parameter for all models. We use greedy sampling during generation and the max generation token is set to 2048. For close-source models, we set $\text{tool\_choice}="auto"$ to let the LLM decide which tool to choose.

\subsection{Golden Function Call Updating}
\label{golden_updating}
\begin{figure}[h]
\centering 
\includegraphics[width=0.5\textwidth]{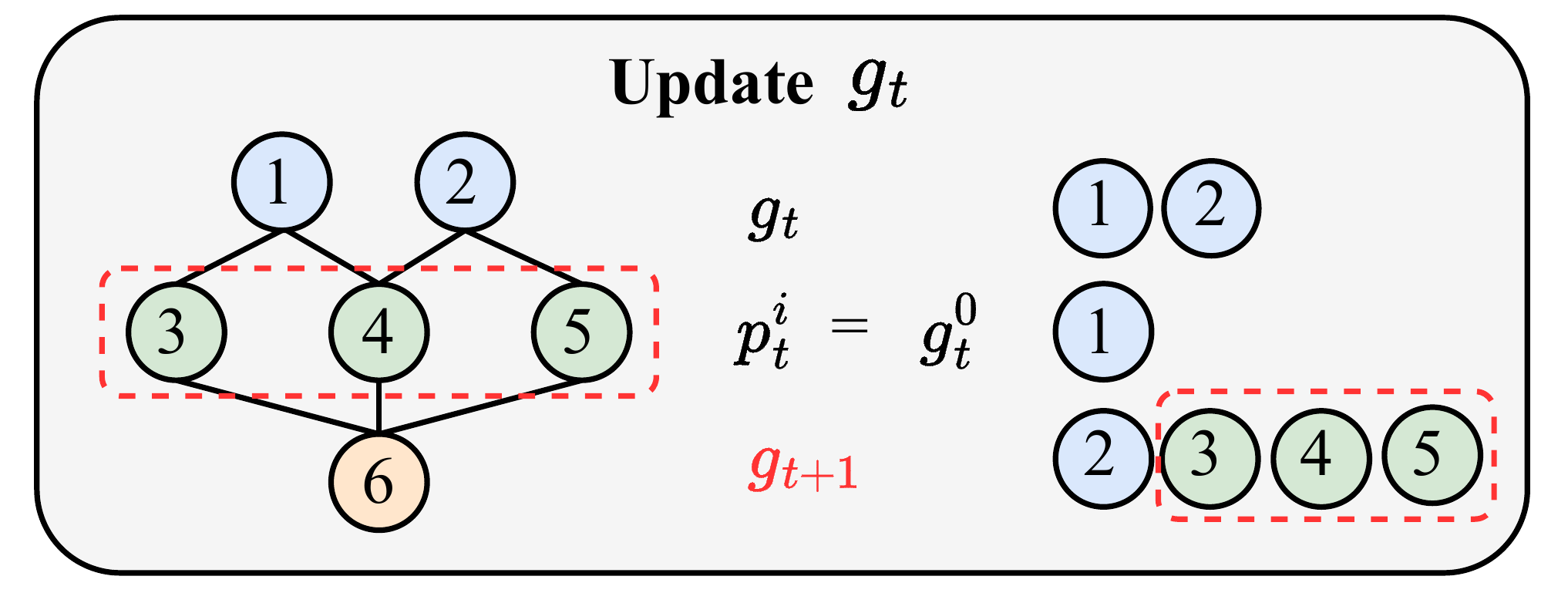}
\vspace{-0.1cm}
\caption{An example for golden function call updating. Path on the left is the annotated shortest function call path with three steps.}
\label{golden_update}
\vspace{-0.2cm}
\end{figure}

As shown in Algorithm \ref{alg1}, the model predicts a function call list with $n$ function calls at step $t$, denoted as $p_t = [p_{t}^{0}, \cdots, p_{t}^{n}]$, while the golden function call list contains $m$ function calls $g_t = [g_{t}^{0}, \cdots, g_{t}^{m}]$. We aim to evaluate the correctness of $p_{t}^{i}$ based on $g_t$. If Any $p_t^i$ matched with $g_t^j$ under the multi-dimensional matching method, the annotated API response of  $g_t^j$ is used as the API response of $p_t^i$, and we will update the golden function call list by adding the next-step golden function calls. Figure \ref{golden_update} is an example if golden function call updating where $t=0$.

\begin{algorithm*}
	\renewcommand{\algorithmicrequire}{\textbf{Input:}}
	\renewcommand{\algorithmicensure}{\textbf{Output:}}
	\caption{Golden Function Call List Updating Strategy}
	\label{alg1}
	\begin{algorithmic}[1]
            \STATE Annotated shortest function call path: $\textbf{G} = [[g_0^1, \cdots, g_0^{m_1}],\cdots,[g_s^1, \cdots, g_s^{m_s}]]$
		\STATE Initialization: $ind \leftarrow 0, t \leftarrow 0$
            \STATE Golden function calls at current step $t$: $ g_t = \textbf{G}[ind] = \textbf{G}[0] = [g_0^1, \cdots, g_0^{m_1}]$
		\REPEAT
            \STATE Model predict function calls at current step $t$:  $p_t = [p_{t}^{0}, \cdots, p_{t}^{n}]$
            \STATE Evaluation $p_t$ based on $g_t$ based on Section \ref{fc_eval}
            \STATE Obtain the success matched golden function calls list $g_t^{success} = [g_t^j, \cdots, g_t^k]$
            \STATE Remain golden function calls $g_t' = g_t - g_t^{success}$
            \IF{$ind < s$}
              \STATE $ind \leftarrow ind + 1$
              \STATE Update Golden function calls $g_{t+1} = g_t' + \textbf{G}[ind]$
            \ELSE
              \STATE Update Golden function calls $g_{t+1} = g_t'$
            \ENDIF
            \STATE $t \leftarrow t + 1$
		\UNTIL Model generate the final response
	\end{algorithmic}  
\end{algorithm*}

\subsection{Prompt for LLM-based Match}
\label{subsec:llm-based-match}
\begin{figure*} %H为当前位置，!htb为忽略美学标准，htbp为浮动图形
\centering %图片居中
\includegraphics[width=0.9\textwidth]{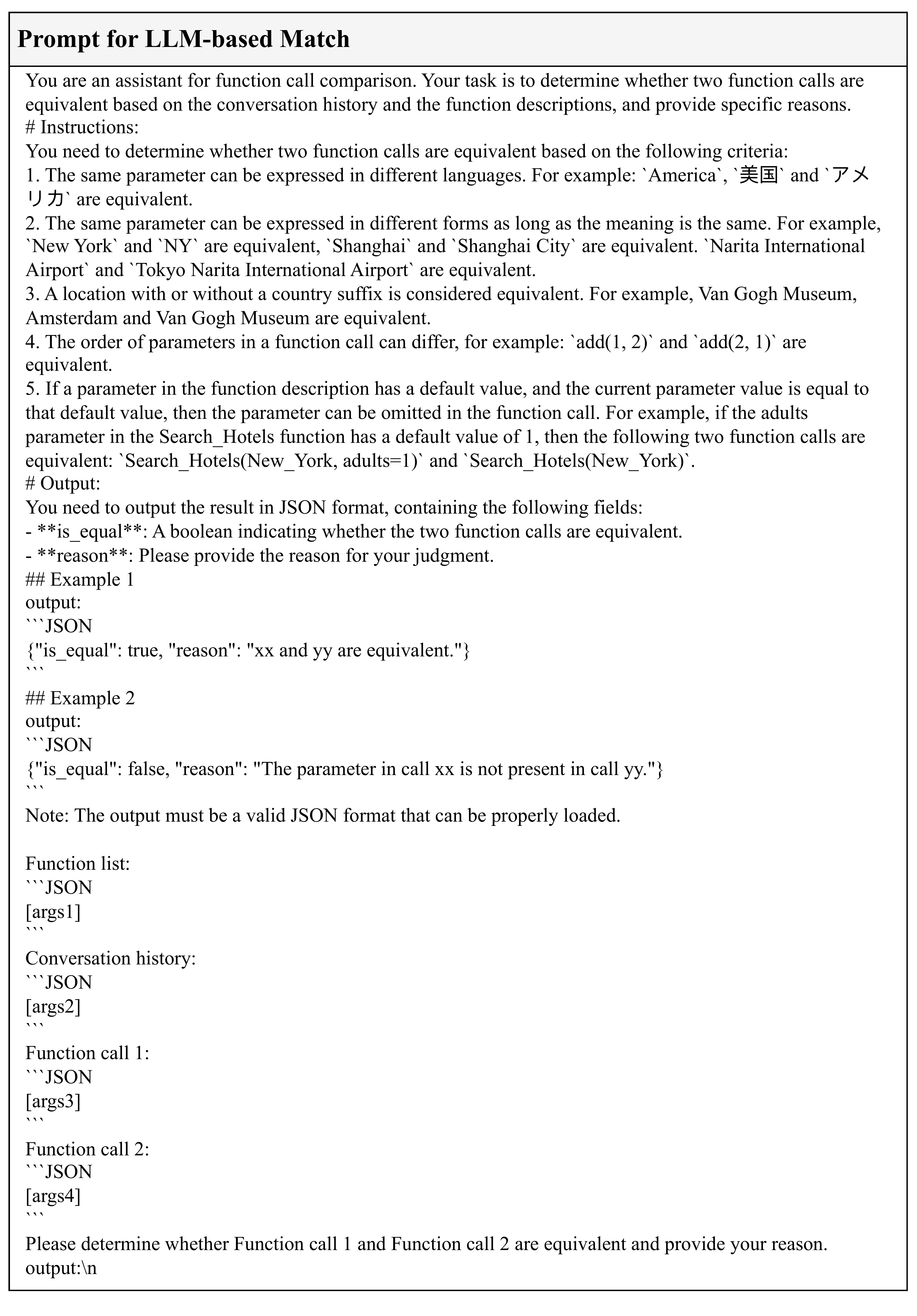}
%插入图片，[]中设置图片大小，{}中是图片文件名
\vspace{-0.1cm}
\caption{Prompt for LLM-based Match.}
\label{llm-based-match}
\vspace{-0.2cm}
\end{figure*}

Prompt for llm-based match is shown in Figure \ref{llm-based-match}.

\subsection{Prompt for Model Response Evaluation}
\label{response_eval_prompt}
\begin{figure*} %H为当前位置，!htb为忽略美学标准，htbp为浮动图形
\centering %图片居中
\includegraphics[width=0.9\textwidth]{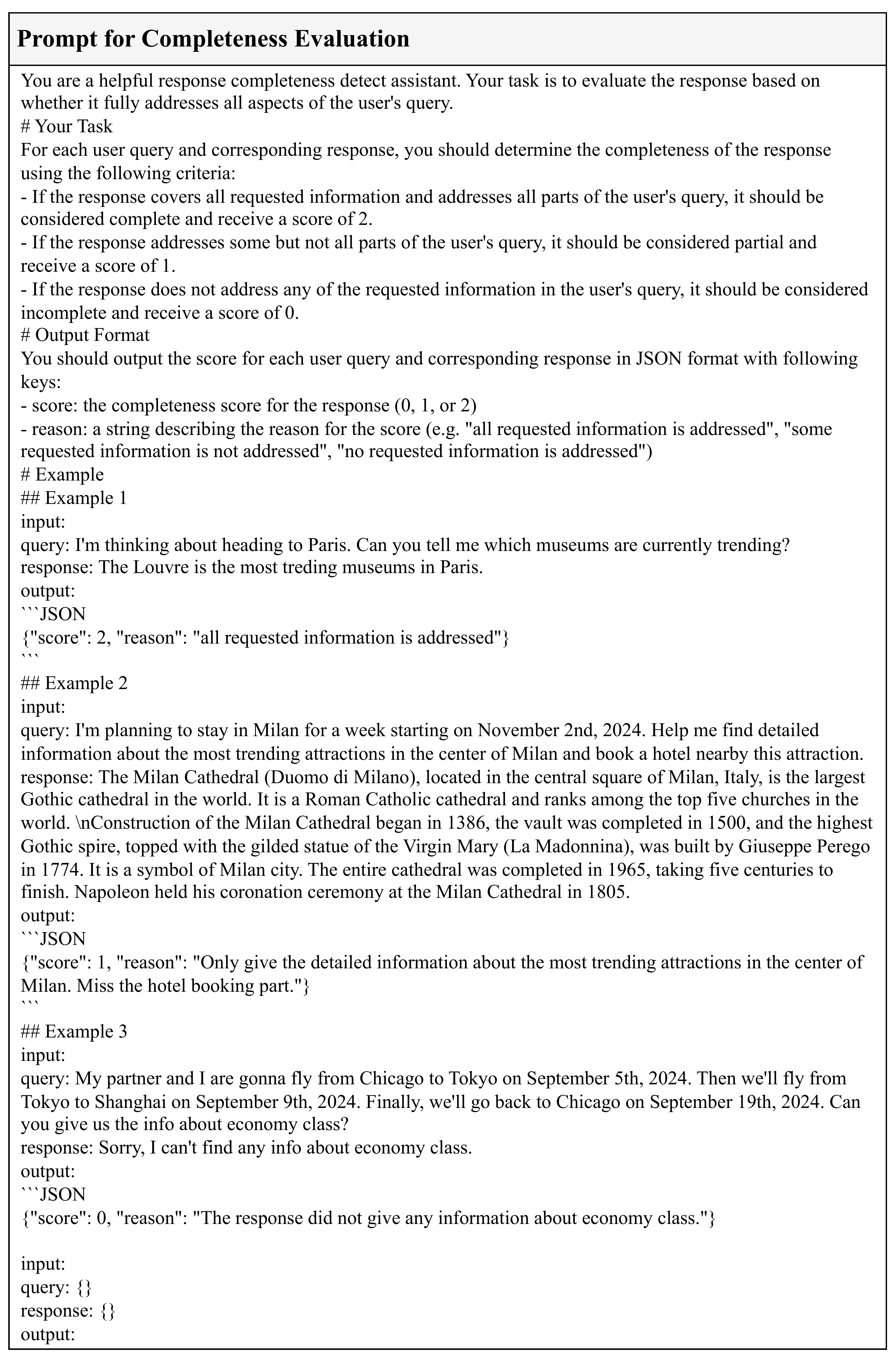}
%插入图片，[]中设置图片大小，{}中是图片文件名
\vspace{-0.1cm}
\caption{Prompt for Completeness Evaluation.}
\label{completeness}
\vspace{-0.2cm}
\end{figure*}

\begin{figure*} %H为当前位置，!htb为忽略美学标准，htbp为浮动图形
\centering %图片居中
\includegraphics[width=0.9\textwidth]{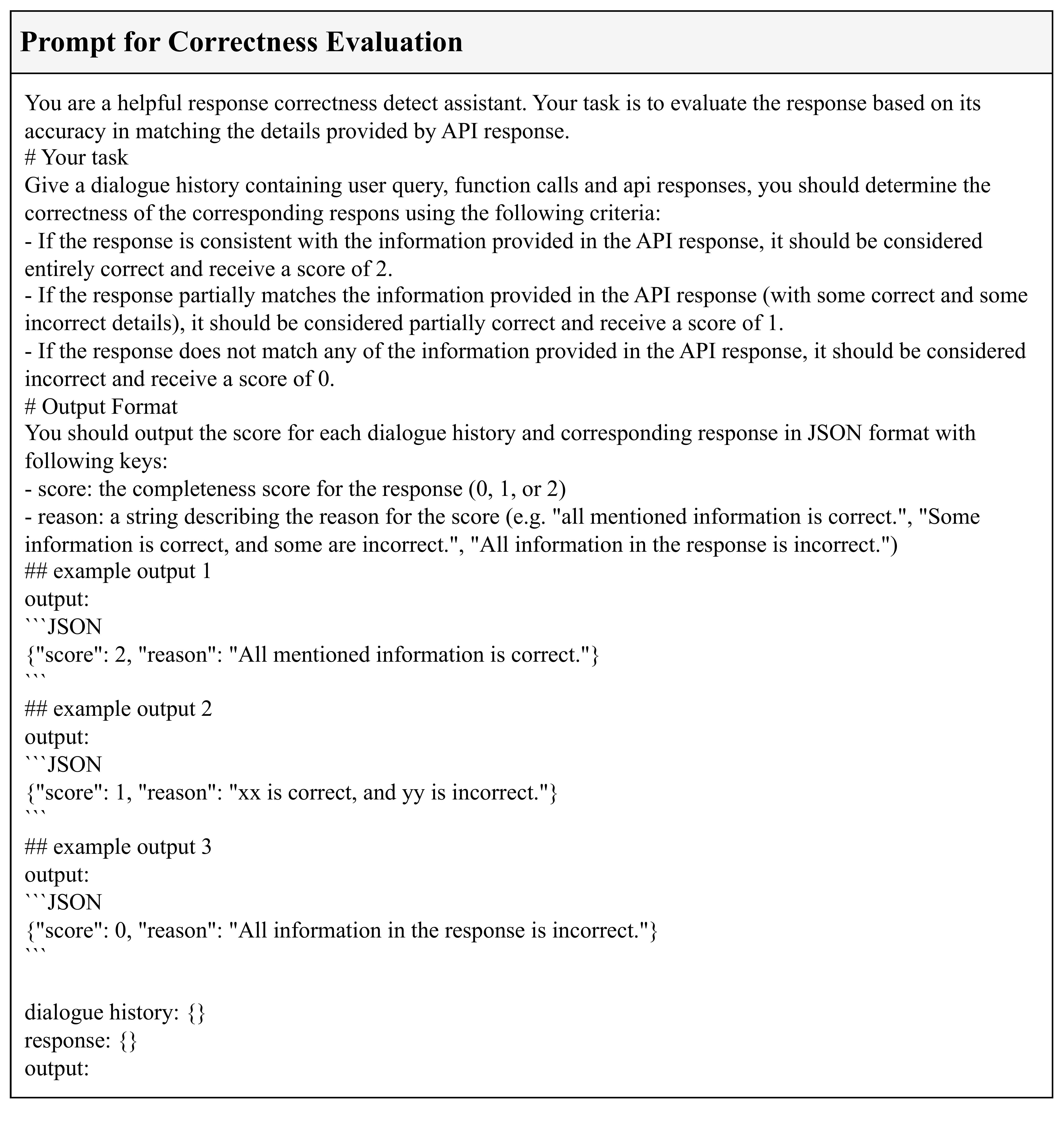}
%插入图片，[]中设置图片大小，{}中是图片文件名
\vspace{-0.1cm}
\caption{Prompt for Correctness Evaluation.}
\label{correctness}
\vspace{-0.2cm}
\end{figure*}

Prompt for model response evaluation are shown in Figure \ref{completeness} and Figure \ref{correctness}.

\section{Parameter Type Examples}
\label{appendix:param_type}
Specific examples of each parameter type are shown in Table  \ref{param_type_ref}.

\begin{table*}
    \centering
    \scalebox{0.8}{
    \begin{tabular}{cp{6cm}p{12cm}} \toprule
         Parameter & Example & Explanation\\ \midrule
         filter & filter = "facility::433, facility::107") &  
         The $filter$ parameter can be retrieved from the API response of the \textsc{Get\_Filter} function. It often appears in queries with constraints, like: Find a few hotels with a pool and free wifi. "facility::433" and "facility::107" denote pool and free wifi, respectively.\\ \hline
         legs &  legs=[\{"fromId": "DFW.CITY", "toId": "MUC.AIRPORT", "date": "2024-11-25"\},\{"fromId": "MUC.AIRPORT", "toId": "STO.CITY", "date": "2024-11-26"\}, \{"fromId": "STO.CITY","toId": "DFW.CITY","date": "2024-12-09"\}])& The $legs$ parameter is the multi-stop flights list which contains location and date. \\ \hline
         token &  token="d7699\_H4sIAAAAAAAA\_ ... AAA."& The $token$ parameter can be retrieved from the API response of multiple functions. For example, the \textsc{Search\_Flights} function will return the $token$ for different flights. \\ \hline
         slug & slug="pr7jepixwlvr-private-guided-tour-orsay-museum-rare-languages"& The $slug$ parameter can be retrieved from the API response of \textsc{Search\_Attraction\_Location} function as as 'productSlug' inside 'products' or 'destinations'.\\ \hline
         date & date="2024-11-22" & Parameters related to dates, like: check-in-date, check-out-date,etc. \\ \hline
         location &  location="Amsterdam"& Parameters related to locations, like: country, city ,etc. \\ \hline
         key & key= "eyJkcml2ZXJzQWdlIjozMCwiZ HJvcE...19GRUVTIl19" & The $key$ parameter can be retrieved from the API response of multiple functions. For example,  the \textsc{Search\_Car\_Rentals} function will return the $key$ for different cars.\\ \hline
         id &  id="eyJ1ZmkiOi01NjQwNjR9" & The $id$ parameter can be retrieved from the API response of multiple functions. For example,  the \textsc{Search\_Attraction\_Location} function will return the $id$ for different attractions.\\ \hline
         time &  time="08:00"& Parameters related to time, like: pick-up-time, drop-off-time,etc.\\ \hline
         sort\_by &  sort\_by="popularity"& The $sort\_by$ parameter controls the order in which hotel or flight results are presented. For example, the hotels can be sorted by price or popularity. \\ \hline
         type & type="landmark"& The $type$ parameter is the entity type, like landmark, city, etc.\\ \hline
         age &  age="8"& The $age$ parameter is the age of people. \\ \hline
         people & people=2 & The $people$ parameter is the number of people. \\ \bottomrule
    \end{tabular}
    }
    \caption{Examples of different parameter types.}
    \label{param_type_ref}
\end{table*}

% All Figures

\begin{table*}[htbp]
\small \centering
\setlength\tabcolsep{4pt}
\begin{tabular}{@{}rp{14cm}@{}}
\toprule
\rowcolor{Gainsboro} \multicolumn{2}{l}{\textbf{\normalsize Query}}\\
\textbf{Initial} & My friend is planning a trip between December 15, 2024 and December 20, 2024 and he wants to fly from Sydney to Melbourne. Please help find the best flight options and book a 4\-star hotel near Fitzroy Gardens. They also need to rent a car and a taxi service from Melbourne Airport to the hotel. \\

\specialrule{0em}{1pt}{1pt}
\hdashline
\specialrule{0em}{1pt}{1pt}

\textbf{Human Annotated} & Please help my friend finds the best flight from Sydney to Melbourne on 15 December 2024 and book a hotel within 10km of Fitzroy Gardens, Melbourne \textcolor{darkgreen}{for one night}. Remember to book a taxi to \textcolor{darkgreen}{pick him up from the airport to the hotel an hour after the plane lands.} \\

\specialrule{0em}{1pt}{1pt}
\hdashline
\specialrule{0em}{1pt}{1pt}

\textbf{Explanation} & \textbf{Correction}. Rewrite query for clarity, such as the number of days for the hotel reservation, the departure location, and the time for the taxi. \\

\specialrule{0em}{1pt}{1pt}
\hline
\specialrule{0em}{1pt}{1pt}
\rowcolor{Gainsboro} \multicolumn{2}{l}{\textbf{\normalsize Function call at step 1}}\\
\multirow{4}{*}{\textbf{GPT Generated}}
& 1. \{"name": "Search\_Flight\_Location", "arguments": \{"query": "Sydney"\}\} \\
& 2. \{"name": "Search\_Flight\_Location", "arguments": \{"query": "Melbourne"\}\} \\
& 3. \{"name": "Location\_to\_Lat\_Long", "arguments": \{"query": "Fitzroy Gardens, Melbourne"\}\} \\
& \textcolor{red}{4. \{"name": "Taxi\_Search\_Location", "arguments": \{"query": "Melbourne Airport"\}\}} \\

\specialrule{0em}{1pt}{1pt}
\hdashline
\specialrule{0em}{1pt}{1pt}

\multirow{3}{*}{\textbf{Human Annotated}}
& 1. \{"name": "Search\_Flight\_Location", "arguments": \{"query": "Sydney"\}\} \\
& 2. \{"name": "Search\_Flight\_Location", "arguments": \{"query": "Melbourne"\}\} \\
& 3. \{"name": "Location\_to\_Lat\_Long", "arguments": \{"query": "Fitzroy Gardens, Melbourne"\}\} \\

\specialrule{0em}{1pt}{1pt}
\hdashline
\specialrule{0em}{1pt}{1pt}

\textbf{Explanation} & \textbf{Correction}. The arrive airport for the best flight may not be "Melbourne Airport". The model need to get the arrive airports before use "Taxi\_Search\_Location". \\

\specialrule{0em}{1pt}{1pt}
\hline
\specialrule{0em}{1pt}{1pt}
\rowcolor{Gainsboro} \multicolumn{2}{l}{\textbf{\normalsize API response at step 1}}\\

\multirow{4}{*}{\textbf{Initial}}
& 1. \{"message": "Success", "data": [\{"id": "SYD.AIRPORT", "name": "Sydney Kingsford Smith Airport"\}, \textcolor{blue}{\{"id": "YQY.AIRPORT", "name": "Sydney (Nova Scotia) Airport"\}]\}} \\
& 2. \{"message": "Success", "data": [\{"id": "MEL.CITY", "name": "Melbourne"\}, \textcolor{blue}{\{"id": "MEL.AIRPORT", "name": "Melbourne Airport"\}, \{"id": "AVV.AIRPORT", "name": "Avalon Airport"\}, \{"id": "MEB.AIRPORT", "name": "Essendon Fields Airport"\}, \{"id": "MLB.AIRPORT", "name": "Melbourne International Airport"\}]\}} \\
& 3. \{"message": "Success", "data": [\{"business\_status": "OPERATIONAL", "formatted\_address": "Wellington Parade, East Melbourne VIC 3002, Australia" "geometry": \{"location": \{"lat": -37.81272149999999, "lng": 144.9800679\}\},"name": "Fitzroy Gardens"\}]\} \\
& \textcolor{red}{4. \{"message": "Success", "data": [\{"longitude": 144.849, "city": "Melbourne", "name": "Melbourne Airport", "latitude": -37.6696\}, \{"city": "Melbourne", "longitude": -80.6456, "name": "Melbourne International Airport", "latitude": 28.1025\}]\}} \\

\specialrule{0em}{1pt}{1pt}
\hdashline
\specialrule{0em}{1pt}{1pt}

\multirow{3}{*}{\textbf{Human Annotated}}
& 1. \{"message": "Success", "data": [\{ "id": "SYD.AIRPORT", "name": "Sydney Kingsford Smith Airport"\}]\}, \\
& 2. \{"message": "Success", "data": [\{"id": "MEL.CITY", "name": "Melbourne"\}]\}, \\
& 3. \{"message": "Success", "data": [\{"business\_status": "OPERATIONAL", "formatted\_address": "Wellington Parade, East Melbourne VIC 3002, Australia" "geometry": \{"location": \{"lat": -37.81272149999999, "lng": 144.9800679\}\},"name": "Fitzroy Gardens"\}]\} \\

\specialrule{0em}{1pt}{1pt}
\hdashline
\specialrule{0em}{1pt}{1pt}

\textbf{Explanation} & \textbf{Disambiguation}. Delete API response that may cause ambiguation for subsequent steps.\\

\specialrule{0em}{1pt}{1pt}
\hline
\specialrule{0em}{1pt}{1pt}
\rowcolor{Gainsboro} \multicolumn{2}{l}{\textbf{\normalsize Function call at step 2}}\\
\multirow{5}{*}{\textbf{GPT Generated}}
& 1. \{"name": "Search\_Flights", "arguments": \{"fromId": "SYD.AIRPORT", "toId": "MEL.CITY", "departDate": "2024-12-15", "sort": "BEST"\}\} \\
& \textcolor{red}{2. \{"name": "Search\_Flights", "arguments": \{"fromId": "YQY.AIRPORT", "toId": "MEL.CITY", "departDate": "2024-12-15", "sort": "BEST"\}\}} \\
& \textcolor{red}{3. \{"name": "Search\_Flights", "arguments": \{"fromId": "SYD.AIRPORT", "toId": "AVV.AIRPORT", "departDate": "2024-12-15", "sort": "BEST"\}\}} \\
& \textcolor{red}{4. \{"name": "Search\_Flights", "arguments": \{"fromId": "YQY.AIRPORT", "toId": "AVV.AIRPORT", "departDate": "2024-12-15", "sort": "BEST"\}\}} \\
& 5. \{"name": "Search\_Hotels\_By\_Coordinates", "arguments": \{"latitude": "-37.81272149999999", "longitude": "144.9800679", "arrival\_date": "2024-12-15", "departure\_date": "2024-12-20", "radius": 10\}\} \\

\specialrule{0em}{1pt}{1pt}
\hdashline
\specialrule{0em}{1pt}{1pt}

\multirow{2}{*}{\textbf{Human Annotated}}
& 1. \{"name": "Search\_Flights", "arguments": \{"fromId": "SYD.AIRPORT", "toId": "MEL.CITY", "departDate": "2024-12-15", "sort": "BEST"\}\} \\
& 2. \{"name": "Search\_Hotels\_By\_Coordinates", "arguments": \{"latitude": "-37.81272149999999", "longitude": "144.9800679", "arrival\_date": "2024-12-15", "departure\_date": "2024-12-20", "radius": 10\}\} \\

\specialrule{0em}{1pt}{1pt}
\hdashline
\specialrule{0em}{1pt}{1pt}

\textbf{Explanation} & \textbf{Correction}. After delete ambiguous information in the API response, we need to remove redundant function calls. \\

\bottomrule
\end{tabular}
\caption{Annotation Example. To make the content easy for reading, we have removed most of the content from the API response. The actual API response contains a large amount of information, reaching a context length of 128k.}
\label{tab:annotate_example}
\end{table*}

\end{document}